\pdfoutput=1
\documentclass[conference]{IEEEtran}
\IEEEoverridecommandlockouts
\usepackage{cite}
\usepackage{amsmath,amssymb,amsfonts}
\usepackage{algorithmic}
\usepackage{algorithm}
\usepackage{graphicx}
\usepackage{textcomp}
\usepackage{xcolor}
\usepackage{booktabs}
\usepackage{multirow}
\usepackage{tabularx}
\usepackage{enumitem}
\usepackage[hyphens]{url}
\usepackage[hidelinks]{hyperref}
\usepackage{eso-pic}
\setlist{nosep, leftmargin=14pt}
\def\BibTeX{{\rm B\kern-.05em{\sc i\kern-.025em b}\kern-.08em
    T\kern-.1667em\lower.7ex\hbox{E}\kern-.125emX}}

\begin{document}

\title{Neuro-Duo-MIL: A Dual-Branch DINOv3 Framework for Leakage-Aware Alzheimer's Disease Classification Using 2D MRI and Clinical Data}

\author{
\IEEEauthorblockN{Rafiul Arefeen Islam}
\IEEEauthorblockA{
\textit{Dept. of Computer Science and Engineering} \\
\textit{Islamic University of Technology}\\
Gazipur, Bangladesh \\
iarefeen41@gmail.com}
\and
\IEEEauthorblockN{Rihila Sumayya}
\IEEEauthorblockA{
\textit{Dept. of Computer Science and Engineering} \\
\textit{Islamic University of Technology}\\
Gazipur, Bangladesh \\
rihila@iut-dhaka.edu}
\and
\IEEEauthorblockN{Adib Ahmed}
\IEEEauthorblockA{
\textit{Dept. of Computer Science and Engineering} \\
\textit{Islamic University of Technology}\\
Gazipur, Bangladesh \\
adibahmed@iut-dhaka.edu}
\and
\IEEEauthorblockN{Md. Hasanul Kabir}
\IEEEauthorblockA{
\textit{Dept. of Computer Science and Engineering} \\
\textit{Islamic University of Technology}\\
Gazipur, Bangladesh \\
hasanul@iut-dhaka.edu}
\and
\IEEEauthorblockN{Rafsanjany Kushol}
\IEEEauthorblockA{
\textit{Dept. of Medicine} \\
\textit{University of Alberta}\\
Edmonton, AB, Canada \\
kushol@ualberta.ca}
}

\AddToShipoutPictureBG*{\AtTextLowerLeft{\put(0,-14){\parbox[t]{\textwidth}{\scriptsize
Accepted for publication in the 2026 International Conference on Image and Vision Computing New Zealand (IVCNZ).\\
\copyright~2026 IEEE. Personal use of this material is permitted. Permission from IEEE must be obtained for all other uses, in any current or future media, including reprinting/republishing this material for advertising or promotional purposes, creating new collective works, for resale or redistribution to servers or lists, or reuse of any copyrighted component of this work in other works.}}}}

\maketitle

\begin{abstract}
Alzheimer's Disease (AD) is one of the most challenging neurodegenerative disorders to diagnose early. Most automated methods rely on a single modality, or on full 3D MRI volumes whose computational cost limits large-scale deployment. We present a subject-level multimodal framework combining axial and coronal 2D MRI slices with clinical and cognitive data. Two separate DINOv3 branches extract features from both views, while multi-instance learning and attention mechanisms identify informative slices and fuse imaging with clinical features. Evaluated on 842 baseline ADNI1 participants under a subject-level split that prevents data leakage, the model achieved 94.49\% test accuracy and a macro F1-score of approximately 0.94 for three-class CN-MCI-AD classification, showing that complementary 2D views combined with clinical information give strong subject-level performance without full 3D volumes.
\end{abstract}


\section{Introduction}
\label{sec:intro}

Alzheimer's Disease (AD) is a progressive neurodegenerative disorder and the leading cause of dementia, responsible for up to 60--70\% of cases worldwide \cite{who_dementia2025}. Early diagnosis is crucial, as no curative treatment exists. Convolutional Neural Networks (CNNs) capture localized structural patterns in MRI, but struggle to model the long-range dependencies across brain regions that characterise global neurodegeneration \cite{Wen2020Convolutional}. Vision Transformers (ViTs) \cite{dosovitskiy2021an} address this through self-attention, yet their data and computational demands limit their use on small medical datasets.

We propose a multimodal framework that integrates 2D MRI slices with clinical features. Using self-supervised visual features from 2D neuroimaging rather than dense 3D volumes substantially reduces computational overhead while preserving diagnostically vital structural biomarkers. Our contributions are threefold: (1) a pipeline built on self-supervised DINOv3 representations for multi-class neuroimaging classification, (2) a slice selection strategy that extracts important diagnostic features from key anatomical views, and (3) an empirical evaluation demonstrating high accuracy and stability under a strict subject-level protocol engineered to ensure zero data leakage.\footnote{Source code: \url{https://github.com/rafiul-arefeen/neuro-duo-mil-alzheimers}}

\section{Related Works}
\label{sec:relatedworks}
The field of automated Alzheimer’s disease classification has evolved rapidly over recent years, driven by advances in deep learning and multimodal representation learning. Existing approaches can broadly be categorized into three directions: CNN-based MRI classification, hybrid CNN--Transformer models, and multimodal fusion frameworks. This section reviews key contributions in each of these areas.

\subsection{CNN-Based MRI Classification Approaches}

Early work in Alzheimer’s disease classification primarily relied on convolutional neural networks (CNNs) applied to structural MRI. These models leveraged spatial patterns such as hippocampal atrophy and cortical thinning to distinguish between cognitively normal (CN) and Alzheimer’s disease (AD) subjects.

Rahman et al. (2025) \cite{rahman2025alzheimer} proposed a 3D-CNN architecture using dilated convolutions and informative frame selection to preserve spatial context while reducing computational cost. Their approach demonstrated strong performance on ADNI by effectively capturing volumetric information from MRI data.

Katkam et al. (2025) \cite{katkam2025integrated} extended this idea by incorporating lesion-aware segmentation into the classification pipeline. Their model combined DeepLabV3+ for segmentation with a classification network, enabling improved localization of disease-relevant regions.

Similarly, Mmadumbu et al. (2025) \cite{mmadumbu2025early} explored transfer learning using pre-trained CNN architectures such as ResNet50 \cite{he2016resnet} and MobileNetV2. Their results showed that deep CNN models can effectively extract features from MRI data, although performance varies depending on architecture choice.

\subsection{Transformer-Based and Hybrid CNN--Transformer Models}

To address the limitation of CNNs in capturing long-range dependencies, recent works have incorporated transformer-based architectures into medical imaging tasks.

Kushol et al. (2022) \cite{Kushol2022} proposed ADDFormer, which fuses spatial-domain features from a ViT with frequency-domain features from a GFNet through a fusion transformer, using selected 2D coronal slices and subject-level majority voting, and reached 88.2\% accuracy on ADNI. Yan et al. (2025) \cite{yan2025hybridrvit} proposed Hybrid-RViT, a model combining ResNet50 with a Vision Transformer. The CNN component extracts local spatial features, while the transformer captures global contextual relationships across brain regions.

Wang et al. (2025) \cite{wang2025cnn} further improved this approach using Swin Transformer, which introduces hierarchical attention and shifted windows for efficient multiscale representation learning. Zhou et al. (2025) \cite{zhou2025deep} combined 3D CNNs with Video Swin Transformer to capture both spatial and temporal dependencies in MRI data. Their approach highlights the importance of combining volumetric modeling with global attention mechanisms.

These hybrid architectures show improved performance over traditional CNNs by integrating both local and global feature representations. However, they still rely primarily on imaging data and do not fully incorporate complementary clinical information.

\subsection{Multimodal Fusion Approaches}

Recent research has increasingly focused on combining multiple data modalities to improve classification performance.

The efficiency of cross-modal attention was demonstrated by Golovanevsky et al. (2022) \cite{golovanevsky2022multimodal}, who enabled effective interaction between MRI, clinical, and genetic features.

To address the challenge of incomplete datasets, Liu et al. (2023) \cite{liu2023cascaded} introduced the 3MT model, which incorporates modality dropout. This specific technique improves robustness in real-world scenarios where complete data is often unavailable. A separate architectural approach is seen in TriFormer (Liu et al., 2023) \cite{liu2023triformer}, which utilizes a transformer-based design with dedicated branches for MRI and clinical data followed by a fusion module.

Bekhet et al. (2025) \cite{bekhet2025deepalznet} introduced DeepALZNET, which uses parallel pathways for MRI and clinical features. Jia et al. (2024) \cite{jia2024globallocal} proposed a global-local fusion framework that combines regional MRI features with clinical patterns.

As deep learning models improved, concerns regarding evaluation methodology became increasingly important. Several studies highlighted the issue of data leakage in MRI-based classification.

Yagis et al. (2021) \cite{yagis2021effect} experimentally demonstrated that slice-level splitting can significantly inflate performance. By comparing different splitting strategies, they showed that models could achieve high accuracy even with randomized labels when leakage is present. Adeoye et al. (2025) \cite{diagnostics15182348} extended this analysis through a large-scale review, confirming that models with strict evaluation protocols report lower but more realistic performance.

These studies highlight the importance of methodological rigor in Alzheimer’s disease classification and emphasize the need for reliable evaluation strategies.

\begin{figure*}[t]
    \centering
    \includegraphics[width=\textwidth]{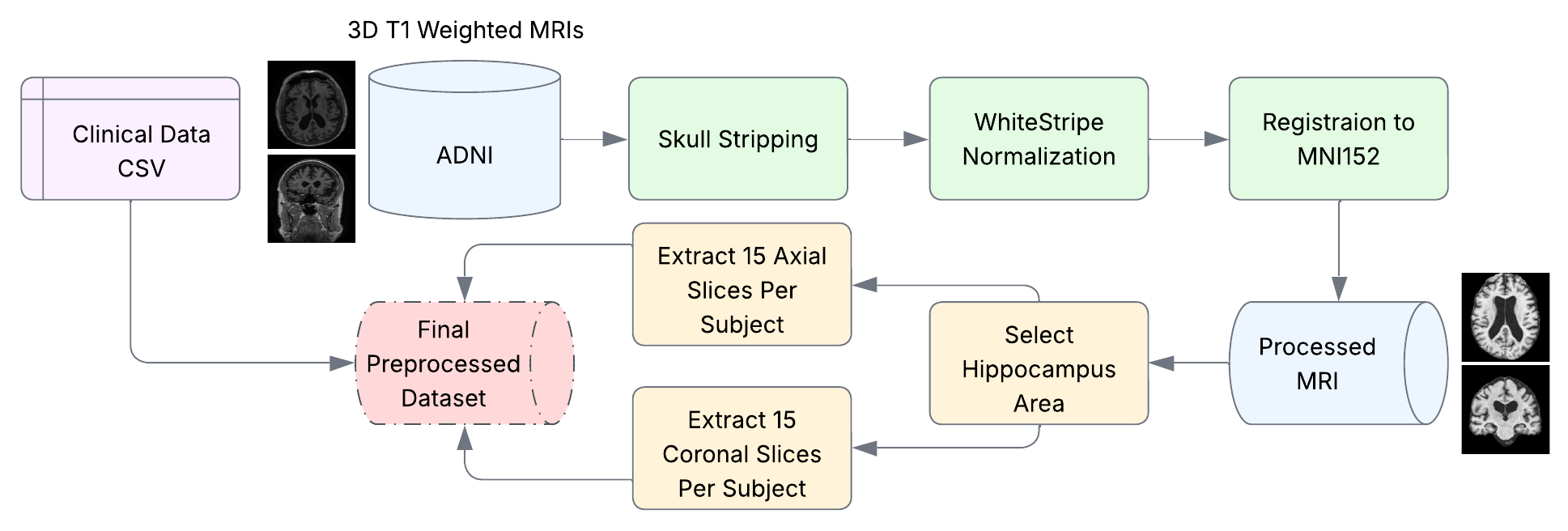}
    \caption{Data preprocessing pipeline}
    \label{fig:preprocessingpipeline}
\end{figure*}

\section{Methods}
\label{sec:methods}
\subsection{Data Preprocessing}

Data used in the preparation of this article were obtained from the Alzheimer’s Disease Neuroimaging
Initiative (ADNI) database \cite{adni} provided by USC Laboratory of Neuroimaging. We used 842 patients from ADNI1, taking only the baseline T1-weighted MRI for each; the distribution is given in Table~\ref{tab:demographics}. AD, CN, and MCI denote Alzheimer's disease, cognitively normal, and mild cognitive impairment, respectively. 
\begin{table}[h!]
\centering
\caption{ADNI1 patient distribution by age range and group.}
\begin{tabular}{lcccccc}
\hline
Group & 50-59 & 60-69 & 70-79 & 80-89 & 90+ & Total \\
\hline
AD   & 7   & 33  & 89  & 70  & 1  & 200 \\
CN   & 0   & 6   & 173 & 51  & 1  & 231 \\
MCI  & 13  & 80  & 197 & 120 & 1  & 411 \\
\hline
\textbf{Total}  &  &  &  &  &  & \textbf{842} \\
\hline
\end{tabular}
\label{tab:demographics}
\end{table}

To account for scanner bias, we applied, in order: skull-stripping using SynthStrip \cite{synthstrip}, WhiteStripe normalization, and registration to the MNI152 template using ANTsPy \cite{antsx}. Note that \textit{patient level splitting} was performed at 70/15/15 ratio prior to slicing, so no patient's slices appeared in both training and testing, ensuring no data leakage. The preprocessing pipeline is shown in Fig.~\ref{fig:preprocessingpipeline}.

We adopted 12 clinical features from Liu et al. \cite{liu2023cascaded}, covering demographic, genetic, cognitive, and functional measures. These are age, gender, education years, APOE4 genotyping, CDR Sum of Boxes (CDRSB), ADAS11, ADAS13, MMSE, and RAVLT immediate, learning, forgetting and percent forgetting. Missing values were handled before analysis and all features standardised with StandardScaler; the scaler fitted on the training data was applied unchanged to validation and test sets.

\subsection{Slice Selection}
To reduce computational cost while preserving diagnostic information, we selected 30 representative 2D slices per volume: 15 axial and 15 coronal. Since the hippocampus is among the earliest and most affected regions in AD \cite{Rao2022Hippocampus}, slices were centred on the hippocampus and surrounding medial temporal structures; the coronal view represents hippocampal atrophy more clearly, while the axial view captures complementary patterns across the brain. Rather than stacking consecutive slices as image channels, each slice was processed independently and the slices of each view treated as instances in separate multi-instance learning bags, letting the model weight the most informative slices. Slices were resized to 224$\times$224. Light augmentation (random horizontal flipping, rotations up to $10^\circ$, minor brightness and contrast jitter) was applied during training.

\subsection{Model Architecture}

\begin{figure*}[t]
    \centering
    \includegraphics[width=1\textwidth]{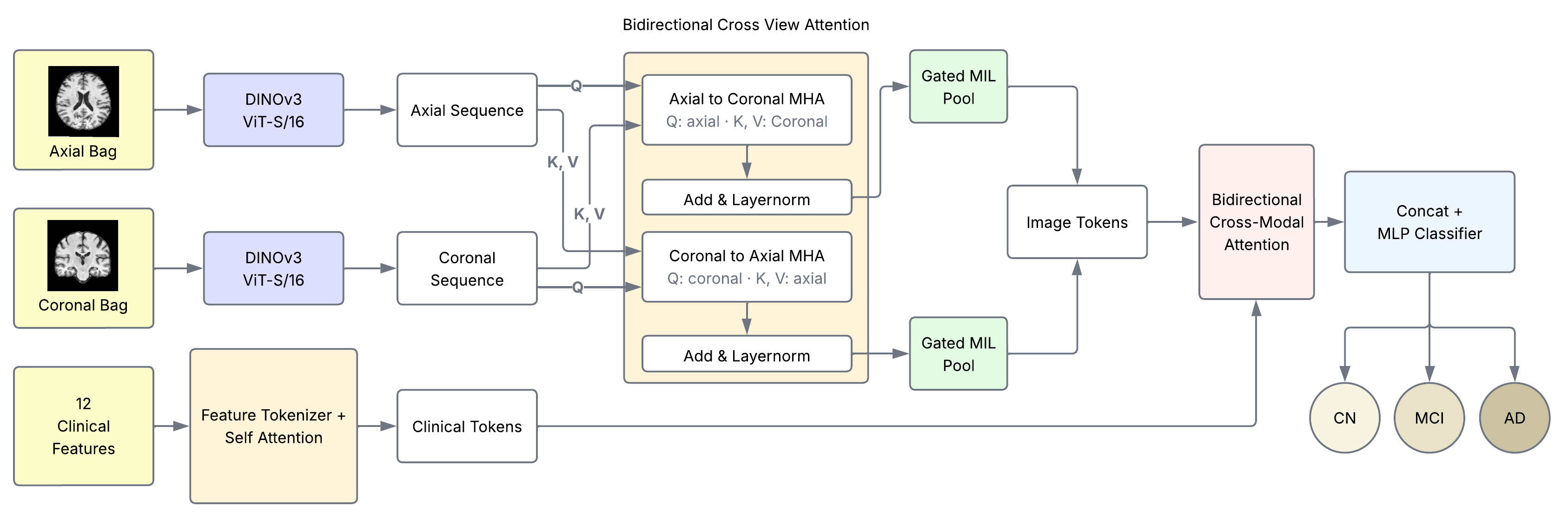}
    \caption{Overview of the proposed dual-view MIL architecture. Per-slice DINOv3 embeddings from the axial and coronal branches are fused with cross-view attention before pooling, then fused again with clinical tokens before classification. MHA stands for Multi-Head Attention.}
    \label{fig:architecture}
\end{figure*}

This section describes the proposed dual-view multiple instance learning
(MIL) architecture from paired axial and coronal MRI slice stacks together with
clinical covariates. We first define the problem
(Sec.~\ref{sec:formulation}), then describe the model from the
slice-level features up: the DINOv3 backbone (Sec.~\ref{sec:backbone}),
cross-view attention (Sec.~\ref{sec:crossview}), gated-attention MIL
pooling (Sec.~\ref{sec:pooling}), clinical feature encoding
(Sec.~\ref{sec:clinical}), multimodal fusion (Sec.~\ref{sec:fusion}), and
the classification head (Sec.~\ref{sec:head}). Sec.~\ref{sec:training}
covers training, including a progressive unfreezing schedule that adapts
the pretrained backbone to a small labeled cohort without destabilizing
it.

\subsubsection{Problem Formulation}
\label{sec:formulation}

Each subject contributes three inputs: a bag of axial slices, a bag of
coronal slices, and a clinical feature vector (up to twelve values,
covering demographics, APOE4 status, and neuropsychological scores such
as MMSE and CDR sum-of-boxes). Slices are resized to $224\times224$ and
bags are capped at 15 slices, sub-sampled uniformly when more are
available. Only one diagnostic label is given per subject, not per
slice, which is what makes this an MIL problem: the model has to learn on
its own which slices, and which correspondences between the two views,
actually matter for diagnosis.

\subsubsection{Architecture Overview}
\label{sec:overview}

Fig.~\ref{fig:architecture} shows the overall design. Two independently
parameterized DINOv3 branches (Sec.~\ref{sec:backbone}), one per view,
embed every slice in their respective bags. The two resulting slice
sequences are fused with cross-view attention (Sec.~\ref{sec:crossview})
before pooling, then each fused sequence is reduced to one vector per
view by gated-attention MIL pooling (Sec.~\ref{sec:pooling}). In
parallel, the clinical vector is tokenized and passed through
self-attention (Sec.~\ref{sec:clinical}). The pooled view embeddings and
clinical tokens are fused via cross-modal attention
(Sec.~\ref{sec:fusion}), and the combined representation is classified by
an MLP (Sec.~\ref{sec:head}).

\subsubsection{Slice-Level Feature Extraction: DINOv3 Backbone}
\label{sec:backbone}

Each view is processed by its own DINOv3 ViT-S/16 encoder
\cite{simeoni2025dinov3} (patch size 16, hidden width 384, twelve
blocks), pretrained on the \texttt{dinov3-vits16-pretrain-lvd1689m}
checkpoint. The two branches are not weight-shared, since the two
anatomical planes look somewhat different. For each slice, the
class-token output is projected into a shared 256-dimensional embedding
space with a linear layer, layer normalization, and a GELU activation.
Applying this to every slice in a bag gives one embedding sequence per
view. Unlike a typical MIL pipeline, this sequence is not pooled yet; it
is passed forward intact so cross-view attention can operate on the full
set of slice tokens.

\subsubsection{Cross-View Attention}
\label{sec:crossview}

Most dual-branch MIL designs pool each view before the two views ever
interact, which throws away information that could be shared between the
axial and coronal planes. We instead fuse the two slice sequences with
bidirectional multi-head attention \cite{vaswani2017attention} before
pooling: axial slices attend over coronal slices, and coronal slices
attend over axial slices, each with a residual connection and layer
normalization.

Doing this before pooling matters, since each bag typically contains
several slices, so the attention softmax has more than one key to weigh
in either direction. We use four attention heads per direction.

\subsubsection{Gated-Attention MIL Pooling}
\label{sec:pooling}

Each cross-view-enhanced sequence is reduced to one subject-level vector
using the gated attention pooling of Ilse et al.
\cite{ilse2018attention}. Each slice embedding passes through a small
gated network (a tanh branch and a sigmoid gate, 128 hidden units) to get
an attention score; scores are normalized with a softmax across the bag,
and the pooled vector is the resulting weighted sum of slice embeddings.

\subsubsection{Tokenized Clinical Feature Encoding}
\label{sec:clinical}

Clinical covariates are standardized using training-set statistics and
clipped at five standard deviations. Instead of concatenating them into
one vector, we tokenize each feature separately, in the spirit of the
FT-Transformer: every scalar feature gets
its own 256-dimensional token through a feature-specific linear
transform. A learnable class token is prepended, and the resulting
sequence is refined by one self-attention block plus a feed-forward
layer, each with a residual connection and layer normalization.

Giving each covariate its own token, rather than packing all of them into
one vector, lets self-attention model interactions between clinical
measures, for example between MMSE and CDR sum-of-boxes, and gives the
fusion stage below a proper multi-token sequence to work with.

\subsubsection{Multimodal Fusion}
\label{sec:fusion}

The two pooled view embeddings form a two-token image sequence, fused
with the clinical token sequence through bidirectional cross-attention,
the same pattern as Sec.~\ref{sec:crossview}: image tokens attend over
clinical tokens, and clinical tokens attend over image tokens, with
residual connections and layer normalization on both sides. The fused
image tokens are averaged into a single summary vector, and the enhanced
clinical class token serves as the clinical-side summary.

\subsubsection{Classification Head}
\label{sec:head}

The pooled axial embedding, pooled coronal embedding, image-fusion
summary, and clinical-fusion summary are concatenated and passed through
a two-layer MLP (GELU activation, dropout 0.2) to produce logits over
CN, MCI, and AD, converted to probabilities with a softmax.

\subsection{Training Procedure}
\label{sec:training}

Table~\ref{tab:training} summarizes the training configuration. Both
backbones start frozen, which lets the randomly initialized pooling,
attention and classifier layers settle on fixed DINOv3 features before more
of the backbone adapts; the last two blocks are then unfrozen, followed by
the full backbone.

\begin{table}[htbp]
\centering
\caption{Training configuration.}
\label{tab:training}
\small
\begin{tabular}{@{}p{0.27\columnwidth}p{0.67\columnwidth}@{}}
\toprule
\textbf{Component} & \textbf{Setting} \\
\midrule
Loss & Class-weighted CE, label smoothing 0.05 \\
Optimizer & AdamW, weight decay $10^{-4}$, gradient clipping at 1.0 \\
LR schedule & One-cycle, up to 70 epochs, batch size 8 \\
Peak LR (heads) & $3\times10^{-4}$ for classifier and fusion modules \\
Peak LR (backbone) & $3\times10^{-5}$, decayed by a factor of 0.75 per transformer block with distance from the output \\
Dropout & 0.2 throughout \\
Unfreezing & Backbones frozen initially; last two blocks at epoch 10, full backbone at epoch 35 \\
Trainable params & 45.4M (full backbone unfrozen) \\
Augmentation & Per subject: same transform shared by all slices in a bag \\
Early stopping & Patience of 20 epochs on validation accuracy, not before epoch 45 \\
Hardware & NVIDIA RTX 4060 8\,GB, bfloat16 mixed precision, gradient checkpointing \\
\bottomrule
\end{tabular}
\end{table}

\section{Results}
\subsection{Experimental Results on the ADNI Dataset}
\label{sec:results}
Classification performance across the three diagnostic cohorts is summarized in Table~\ref{tab:classification_report}.

\begin{table}[htbp]
\centering
\caption{Classification report on the test set}
\label{tab:classification_report}
\begin{tabular}{lcccc}
\hline
\textbf{Class} & \textbf{Precision} & \textbf{Recall} & \textbf{F1-score} & \textbf{Support} \\
\hline
CN  & 1.00 & 0.94 & 0.97 & 35 \\
MCI & 0.94 & 0.95 & 0.94 & 62 \\
AD  & 0.90 & 0.93 & 0.92 & 30 \\
\hline
\textbf{Accuracy} &  &  & 0.94 & 127 \\
\textbf{Macro Avg} & 0.95 & 0.94 & 0.94 & 127 \\
\textbf{Weighted Avg} & 0.95 & 0.94 & 0.95 & 127 \\
\hline
\end{tabular}
\end{table}

\begin{table*}[t]
\centering
\caption{Comparison with representative Alzheimer's disease classification methods.}
\label{tab:sota_comparison}
\begin{tabular}{lllll}
\toprule
\textbf{Model} & \textbf{Dataset} & \textbf{Class} & \textbf{Accuracy (\%)} & \textbf{Evaluation Protocol} \\
\midrule
3MT~\cite{liu2023cascaded}
    & ADNI1 & MCI vs AD & 76.73 & Subject-level \\
    & ADNI2 & MCI vs AD & 83.33 & Subject-level \\
\addlinespace

Novel CNN~\cite{elassy2024novel}
    & ADNI & 3-Class & 99.43 & Image-level split$^{\ast}$ \\
\addlinespace

Hybrid-RViT~\cite{yan2025hybridrvit}
    & OASIS & 4-Class & 95.00 & Subject-level integrity not reported$^{\dagger}$ \\
\addlinespace

CNN Ensemble~\cite{shahid2025novel}
    & ADNI & 4-Class & 96.96 & Augmentation before split$^{\ddagger}$ \\
    & OASIS & 4-Class & 98.35 & Augmentation before split$^{\ddagger}$ \\
\addlinespace

MADDi~\cite{golovanevsky2022multimodal}
    & ADNI & 3-Class & 96.88 (70\% MCI Recall) & Subject-level integrity not reported$^{\dagger}$ \\
\addlinespace

3D-CNN-VSwin~\cite{zhou2025deep}
    & ADNI & CN vs AD & 92.92 & Subject-level \\
\addlinespace

ECAResNet269~\cite{bayahya2025comparative}
    & ADNI1 & 3-Class & 63.00 (72\% MCI Sensitivity) & Subject-level \\
\addlinespace

Automated CNN~\cite{basaia2019automated}
    & ADNI & eMCI vs lMCI & 75.00 & Subject-level \\
\midrule

\textbf{Our Model}
    & \textbf{ADNI1} & \textbf{3-Class} & \textbf{94.49} & \textbf{Subject-level} \\
\bottomrule
\end{tabular}

\vspace{1mm}
\footnotesize
$^{\ast}$Volumes were sliced and then split, allowing scans from one volume to appear in different splits.
\\
$^{\dagger}$Subject-level separation is not explicitly verified, permitting leakage if scans or slices from the same subject appear across splits. 
\\
$^{\ddagger}$Augmentation was performed before splitting, allowing augmented versions of the same image to appear in different splits.
\end{table*}

The model achieved 94\% overall accuracy, isolating the control cohort with perfect precision (1.00) while maintaining balanced sensitivity and specificity across the harder clinical transitions (F1 of 0.94 for MCI and 0.92 for AD). Macro and weighted averages remained between 0.94 and 0.95 across all metrics, confirming that performance is stable and not skewed by the class distribution within the 127 evaluation subjects.

\begin{figure}[htbp]
    \centering
    \includegraphics[width=1\linewidth]{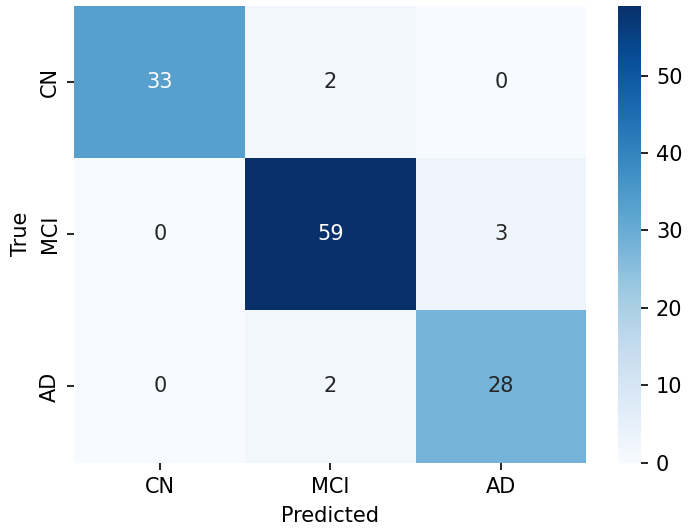}
    \caption{Confusion Matrix for CN vs AD vs MCI Classification}
    \label{fig:confusion}
\end{figure}

The confusion matrix in Fig.~\ref{fig:confusion} shows strong precision along the main diagonal: 33 of 35 CN, 59 of 62 MCI and 28 of 30 AD subjects are classified correctly. Crucially, there are no misclassifications between the polar extremes (CN and AD); the minimal off-diagonal errors occur strictly between adjacent clinical stages, supporting the model's reliability in learning progressive pathological trajectories.

\subsection{Comparison to Other Models}

Table~\ref{tab:sota_comparison} compares our method with representative approaches. Several report higher accuracies, but these must be read together with their evaluation protocols: some split at the image level or augment before partitioning, allowing highly correlated samples from the same subject to appear in both training and test sets, while others do not explicitly verify subject-level separation. Our method instead uses one scan per subject. Thus each subject is assigned exclusively to the training, validation or test set. This prevents patient overlap and gives a more realistic assessment of generalization to unseen subjects; although more challenging and numerically less flattering than leakage-prone protocols, it better reflects clinical deployment. Cross-study accuracy comparisons should therefore be interpreted with caution.

\section{Ablation Study}
To validate the choice of feature extractor, we evaluated the framework across a range of convolutional and transformer-based backbones. 

\begin{table}[htbp]
\centering
\caption{Performance comparison of different model backbones}
\label{tab:backbone_comparison}
\begin{tabular}{lcc}
\toprule
\textbf{Backbone} & \textbf{Accuracy (\%)} & \textbf{F1 Macro (\%)} \\
\midrule
\textbf{DINOv3} & \textbf{94.49} & \textbf{94.42} \\
ResNet50 + ViT            & 86.19 & 86.39 \\
ResNet18 + ViT           & 85.55 & 85.74 \\
ConvNeXt Base v1 + ViT    & 85.32 & 85.67 \\
ResNet101 + ViT           & 84.79 & 85.15 \\
DenseNet201 + ViT         & 84.76 & 85.28 \\
EfficientNet B2 + ViT     & 84.25 & 84.51 \\
EfficientNet B4 + ViT     & 83.66 & 84.02 \\
MobileNet V3 Small + ViT  & 81.95 & 82.50 \\
DenseNet169 + ViT         & 81.86 & 82.24 \\
ResNet34 + ViT            & 81.69 & 82.15 \\
MobileNet V3 Large + ViT  & 81.66 & 82.14 \\
DenseNet121 + ViT         & 81.16 & 81.64 \\
EfficientNet B0 + ViT     & 80.91 & 81.33 \\
\bottomrule
\end{tabular}
\end{table}

As summarized in Table~\ref{tab:backbone_comparison}, supervised convolutional models (the ResNet, ConvNeXt and EfficientNet families) consistently struggled to capture the pathological markers required for robust classification, with the best baseline (ResNet50) capping at 86.19\%. DINOv3 yielded a substantial leap to 94.49\% accuracy --- around 8\% over the strongest convolutional alternative --- justifying its selection as our backbone.

\section{Conclusion}
\label{sec:conclusion}
We presented Neuro-Duo-MIL, a computer vision framework for Alzheimer's disease classification from structural neuroimaging and clinical data in ADNI. Leveraging the self-supervised representations of a DINOv3 vision transformer, our model reached 94.49\% test accuracy and 0.94 Macro F1-score on the rigorous three-class CN vs. MCI vs. AD task while strictly ensuring no data leakage. Future work will focus on domain adaptation to mitigate inter-cohort variability, and on extending the framework to longitudinal prediction and early and late-stage MCI progression detection.

\section*{Compliance with Ethical Standards}
\label{sec:ethics}
This research used data obtained from ADNI in accordance with their data use policies. No efforts were made to identify any individual participant.

\section*{Acknowledgment}
Data collection and sharing for the Alzheimer's Disease Neuroimaging Initiative (ADNI) is funded by the National
Institute on Aging (National Institutes of Health Grant U19 AG024904). The authors declare they have no conflicts of interest. 

\bibliographystyle{IEEEtran}
\bibliography{refs}

\end{document}